\documentclass[11pt,a4paper]{article}
\usepackage{natbib}
\usepackage{graphicx}
\usepackage{wrapfig}
\usepackage[labelfont=bf]{caption}
\usepackage{amsmath}
\usepackage{amsfonts}
\usepackage{amssymb}
\usepackage{makeidx}
\usepackage{setspace}
\usepackage{geometry}
\usepackage{color}
\usepackage{hyperref}
\usepackage{xcolor}
\usepackage{epigraph}
\hypersetup{colorlinks, linkcolor={red!50!black}, citecolor={blue!50!black}, urlcolor={blue!80!black}}
\title{Scaling may be all you need for achieving human-level object recognition capacity with human-like visual experience}
\author{
  Emin Orhan \\
  New York University\\
  \texttt{eo41@nyu.edu}
  }
\date{}

\begin{document}

\maketitle

\begin{abstract}
This paper asks whether current self-supervised learning methods, if sufficiently scaled up, would be able to reach human-level visual object recognition capabilities with the same type and amount of visual experience humans learn from. Previous work on this question only considered the scaling of data size. Here, we consider the simultaneous scaling of data size, model size, and image resolution. We perform a scaling experiment with vision transformers up to 633M parameters in size (ViT-H/14) trained with up to 5K hours of human-like video data (long, continuous, mostly egocentric videos) with image resolutions of up to 476$\times$476 pixels. The efficiency of masked autoencoders (MAEs) as a self-supervised learning algorithm makes it possible to run this scaling experiment on an unassuming academic budget. We find that it is feasible to reach human-level object recognition capacity at sub-human scales of model size, data size, and image size, if these factors are scaled up simultaneously. To give a concrete example, we estimate that a 2.5B parameter ViT model trained with 20K hours (2.3 years) of human-like video data with a spatial resolution of 952$\times$952 pixels should be able to reach roughly human-level accuracy on ImageNet. Human-level competence is thus achievable for a fundamental perceptual capability from human-like perceptual experience (human-like in both amount and type) with extremely generic learning algorithms and architectures and without any substantive inductive biases.
\end{abstract}

\section{Introduction}
\epigraph{``I should have liked teaching it tricks very much, if—if I'd only been the right size to do it!''}{---\textit{Alice's Adventures in Wonderland}, \href{https://en.wikisource.org/wiki/Alice's_Adventures_in_Wonderland_(1866)/Chapter_4}{Chapter 4}}
\vspace{-1.4em}
Are modern self-supervised learning (SSL) algorithms as data efficient as humans in learning powerful internal models of the world? Here, we address this question with respect to a specific capability, namely real-world visual object recognition. A direct comparison of the data efficiency of deep learning models trained with SSL algorithms \textit{vs.}~humans is challenging for a number of reasons:
\begin{itemize}
    \item \textbf{Mismatch in training data:} These models are typically trained with a very different kind of visual data from humans, both in terms of amount and type of data.
    \item \textbf{Mismatch in model size:} They are typically much smaller in size than the human brain, or even just the visual areas in the human brain, \textit{e.g.}~comparing the number of parameters in a model with the number of synapses in a human brain.\footnote{A conservative estimate of 10\% of the brain dedicated to visual processing and 10K synapses per neuron suggests an estimate of 100T synapses or connections dedicated to visual processing in the human brain. This number is several orders of magnitude larger than the number of parameters used in current deep learning models.}
    \item \textbf{Mismatch in input size:} They typically operate on much smaller (lower dimensional) inputs than the human brain works with, \textit{e.g.}~comparing typical image sizes used in computer vision with the number of photoreceptors in the human retina.\footnote{The human retina contains roughly 6M color sensitive cone receptors, very tightly concentrated within a few degrees of visual angle around the fovea \citep{williamson1983}. By comparison, the average size of an ImageNet image (after resizing) is 310$\times$256 pixels, or 0.08 megapixels (MP), in the most commonly used preprocessing pipeline today, which is about two orders of magnitude smaller in spatial resolution.}
\end{itemize} 
We perform a scaling experiment to address these mismatches and ask if current SSL algorithms can reach human-level visual object recognition capabilities at sub-human scales of (human-like) data size, model size, and input size. This is a direct extension of our earlier work \citep{orhan2021}, which considered only data scaling (fixing the model size and input size to relatively small values) and concluded that current SSL algorithms may be orders of magnitude less data efficient than humans. Here, we consider the simultaneous scaling of all three factors and reach a different conclusion (we also train our models on a roughly 4$\times$ larger set of human-like videos in the current experiments). Code and models are available from the following public repository: \href{https://github.com/eminorhan/humanlike-vits}{https://github.com/eminorhan/humanlike-vits}. 

\section{Experiments}
\subsection{Training data}

\begin{wrapfigure}[17]{r}{0.43\textwidth}
  \centering
	\includegraphics[width=0.43\textwidth, trim=2mm 2mm 2mm 2mm, clip]{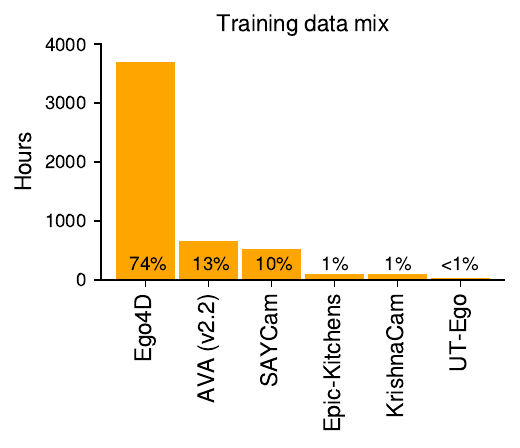} 
	\caption{Human-like video datasets used for SSL.}
  \label{data_mix_fig}
\end{wrapfigure}

The full training data consists of a combination of six video datasets totaling \colorbox{lime}{4971 hours of human-like video}. The videos are ``human-like'' in two prominent aspects: (i) most of the videos are naturalistic, egocentric headcam videos recorded from the perspective of adult or child camera wearers during the course of their daily lives; (ii) they are temporally extended, continuous videos typically lasting tens of minutes to hours long in duration (unlike common video datasets in computer vision, which typically contain much shorter videos). The combined training dataset consists of the following video datasets:

\textbf{Ego4D}: Ego4D is a large dataset of egocentric headcam videos recorded by 923 unique participants while performing daily activities \citep{grauman2022}. The total length of the videos in Ego4D is about 3670 hours. The dataset can be accessed from \href{https://ego4d-data.org/}{this address} after signing a license agreement. 

\textbf{AVA (v2.2)}: AVA is a dataset of films \citep{gu2018}. The dataset comes with rich annotations for parts of these films. However, we use the entire films (without annotations) for self-supervised learning in this work. We use all videos in both \texttt{trainval} and \texttt{test} splits (299 and 131 videos respectively). The total length of the videos from this dataset is about 636 hours. The dataset is publicly accessible from \href{https://research.google.com/ava/download.html}{this address}.

\textbf{SAYCam}: SAYCam is a large, longitudinal dataset of headcam videos recorded from the perspective of three young children between the ages of 6 to 32 months \citep{sullivan2021}. The combined length of the videos from all three children is about 498 hours. Recordings were made approximately once a week for 1-2 hours (usually continuously) over the course of a 2.5 year period during the early development of the children. This dataset is not publicly accessible, but researchers affiliated with an institution can apply for access on the \href{https://databrary.org/about/agreement.html}{Databrary} repository for behavioral science.

\textbf{Epic-Kitchens}: This dataset contains relatively long headcam videos from multiple participants performing daily culinary chores in their kitchens \citep{damen2018}. The total length of the videos in this dataset is about 80 hours. The dataset is publicly accessible from \href{https://epic-kitchens.github.io/2021}{this address}.

\textbf{KrishnaCam}: KrishnaCam is a dataset of headcam videos recorded by a graduate student \citep{singh2016}. It contains long, continuous, egocentric videos of daily episodes in the life of the graduate student. The total length of the videos in the dataset is about 70 hours. The dataset is publicly accessible from \href{https://krsingh.cs.ucdavis.edu/krishna_files/papers/krishnacam/krishnacam.html}{this address}.

\textbf{UT Ego}: UT Ego contains four continuous, egocentric headcam videos, each 3-5 hours long \citep{lee2012} and is publicly accessible from \href{http://vision.cs.utexas.edu/projects/egocentric_data/UT_Egocentric_Dataset.html}{this address}.

The videos were temporally subsampled at the rate of 1 fps, yielding over 17M frames in total for the full dataset.

Since our training data is a combination of multiple video datasets collected from multiple individuals, it does not represent the visual experiences of a single individual, which creates a mismatch between the statistics of our training data and the typical visual experiences of a single person. However, in our recent work \citep{orhan2023}, we found that models trained on length-matched subsets of different video datasets with similar temporal characteristics, \textit{e.g.} SAYCam and Ego4D, perform remarkably similarly in downstream evaluation tasks, including in ImageNet and out-of-distribution versions of ImageNet (two benchmarks for visual object recognition we focus on in this paper), despite notable qualitative differences between these video datasets in terms of both content and style.\footnote{This result is reminiscent of the robust emergence of linguistic capabilities in children despite large differences in the amount and type of linguistic input they receive during their development both across and within cultures \citep{cristia2019,bergelson2019}.} This result suggests that these datasets (and likely other datasets with similar temporal characteristics) may be, by and large, interchangeable with each other insofar as their effect on overall performance in downstream evaluation tasks is concerned. However, this does not necessarily preclude more fine-grained differences between models trained on different datasets or different subsets of the same dataset.

Figure~\ref{data_mix_fig} shows a breakdown of the individual video datasets in our combined training data together with their sizes (in hours). By far the biggest contributor to our combined training data is Ego4D (74\%). The private SAYCam dataset constitutes only about 10\% of the training data. Therefore, we expect that those who wish to replicate the experiments in this work using only publicly accessible data sources should get very similar results to those reported here.

To determine the scaling of object recognition performance with the amount of human-like visual data used during self-supervised pretraining, we train models on the entire dataset and on continuously sampled random subsets of it \citep{orhan2021}. Specifically, we train models on 100\%, 10\%, 1\%, 0.1\%, and 0.01\% of the entire dataset (all sampled as continuous chunks or segments of video). These subsets contain from roughly 5000 to 0.5 hours of human-like video respectively (from the largest to the smallest one), thus covering a \colorbox{lime}{10000-fold range in data size}. Since the subset selection is stochastic, we repeat it three times for each data size. This gives us a total of 13 different datasets (1 entire data + 4 proper subsets $\times$ 3 repeats), on which we train all of our models.     

\subsection{Models}
We exclusively use vision transformer (ViT) models in our experiments \citep{dosovitskiy2020}. We consider four standard sizes for our models: ViT-S, ViT-B, ViT-L, and ViT-H, all with 14$\times$14 patches. These models respectively have 22M, 87M, 304M, and 633M parameters, covering a \colorbox{lime}{29-fold range in model size} from the smallest to the largest model. For the image resolutions, we consider three different spatial resolutions: 224$\times$224 (0.05 MP), 448$\times$448 (0.2 MP), and 476$\times$476 pixels (0.23 MP), covering a roughly \colorbox{lime}{4.5-fold range in spatial resolution} (number of pixels). Due to computational costs, we use the larger two resolutions with the ViT-H model only. This gives us a total of 6 different model architectures (ViT-S/14, ViT-B/14, ViT-L/14, ViT-H/14, ViT-H/14@448, and ViT-H/14@476), which we train on each of the 13 datasets described in the previous subsection for a grand total of 6 $\times$ 13 = 78 pretrained models.

\subsection{SSL algorithm}
We use \colorbox{lime}{masked autoencoders} (MAEs) as our SSL algorithm of choice \citep{he2022}. MAEs have a number of advantages over alternative self-supervised visual representation learning algorithms that are relevant for our purposes in this work. First, unlike most other visual SSL algorithms, they require very minimal data augmentation. \cite{he2022} show that using only random crops works well in MAEs. This is advantageous for our purposes, because heavy data augmentations used in other SSL algorithms make the training data less human-like. Second, MAEs work well with very high masking ratios (we use a masking ratio of 80\%). Because only visible (unmasked) patches are passed through the encoder in MAEs, we can train much bigger models with larger image sizes than we would be able to do with other algorithms on a modest academic compute budget. We use a fairly standard training configuration for MAEs. We refer the reader to the accompanying \href{https://github.com/eminorhan/humanlike-vits}{GitHub repository} for further details. 

\section{Results}
We use validation accuracy on ImageNet \citep{russakovsky2015} and OOD accuracy on out-of-distribution (OOD) versions of ImageNet \citep{geirhos2021} as our proxies for real-world visual object recognition capacity. These benchmarks emphasize both accuracy and robustness of visual object recognition for real-world objects. Both benchmarks come with human performance estimates: for ImageNet, we consider 90\% top-5 accuracy as a reasonable lower-bound on human performance based on the human experiments in \cite{russakovsky2015}; for the OOD versions of ImageNet, the average top-1 accuracy for humans is reported to be 72.3\% in \cite{geirhos2021}. 

\begin{figure}
  \centering
    \includegraphics[width=1.0\textwidth, trim=2mm 2mm 2mm 2mm, clip]{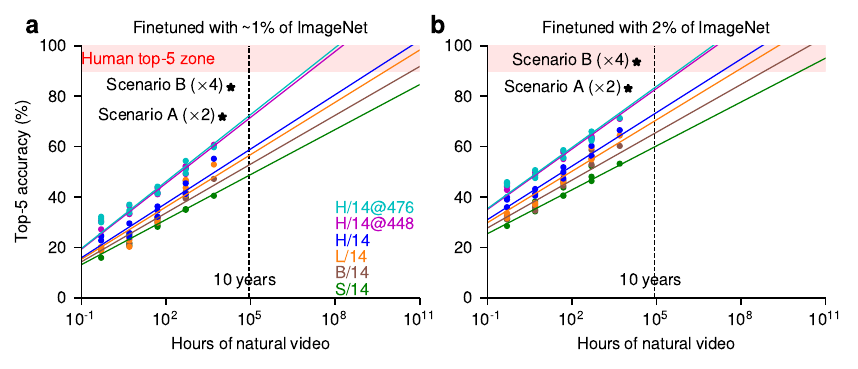}
  \caption{Top-5 validation accuracy on ImageNet as a function of the amount of human-like video data used for self-supervised pretraining. Human-level performance is indicated by the red zone at the top ($>$90\%). For reference, the developmentally relevant time scale of 10 years is indicated by the vertical dashed line. Different colors correspond to six different models identified in the legend. The solid lines represent the fits to Equation~\ref{poly_eq}. Based on these fits, projected accuracies under two hypothetical scenarios are shown by the $\star$ symbols (see Table~\ref{imagenet_tab} for a specification of these scenarios). \textbf{a} shows the results for the stringent finetuning condition (models finetuned with $\sim$1\% of ImageNet) and \textbf{b} shows the results for the more permissive finetuning condition (models finetuned with 2\% of ImageNet).}
  \label{imagenet_fig}
\end{figure}

\begin{table}
\centering
\begin{tabular}{cccccc}
 Scenario & $n$ (years) & $d$ (params) & $r$ (pixels) & Top-5 acc ($\sim$1\%) & Top-5 acc (2\%) \\[1mm]
\hline
 Reference (actual) & 0.6 & 0.6B & 476$\times$476 & 59.8 & 71.4 \\
 Scenario A ($\times$2 projection) & 1.2 & 1.3B & 672$\times$672 & 71.8 & 83.1 \\
 \colorbox{lime}{Scenario B ($\times$4 projection)} & \colorbox{lime}{2.3} & \colorbox{lime}{2.5B} & \colorbox{lime}{952$\times$952} & 83.5 & \colorbox{lime}{93.6} \\
\end{tabular}
\caption{Validation accuracy on ImageNet under an actual reference scenario and two projections based on the fits to Equation~\ref{poly_eq}. The last two columns show the actual and projected accuracy under the stringent ($\sim$1\%) and permissive (2\%) finetuning conditions respectively.}
\label{imagenet_tab}
\vspace{0em}
\end{table}

For both benchmarks, we measure the performance of the pretrained models after few-shot supervised finetuning. We evaluate the models under two different finetuning conditions: (i) a stringent setting that strictly replicates the human experiments in \cite{russakovsky2015} and \cite{geirhos2021}; and (ii) a more permissive setting that uses additional labeled examples for supervised finetuning. The more permissive setting is intended to partially account for any prior (supervised) exposure humans might have had to ImageNet categories before the experiment. In the ImageNet benchmark, the strict setting uses exactly 13000 labeled examples from the ImageNet training set ($\sim$1\% of the training set) for finetuning; the more permissive setting uses 2\% of the training set for finetuning. In the OOD ImageNet benchmark, the strict setting only uses the 321 labeled practice images used by \cite{geirhos2021} to familiarize the human participants with the stimuli and the experiment during an initial practice session; the more permissive setting additionally uses 2\% of the ImageNet training set for supervised finetuning.    

All experimental details related to finetuning and evaluation here are identical to \cite{orhan2021}, so we encourage the reader to refer to \cite{orhan2021} for further details.

Figure~\ref{imagenet_fig} shows the results of our scaling experiment on ImageNet. We model the effects of data size, model size, and image resolution on object recognition accuracy as a simple polynomial function in log space: 
\begin{equation}
    \text{\normalfont accuracy} = {\color{violet}\underbrace{(\alpha_n\log n + \beta_n)}_{\text{\normalfont data scaling}}}  {\color{teal}\underbrace{(\alpha_d \log d + \beta_d)}_{\text{\normalfont model size scaling}}}  {\color{brown}\underbrace{(\alpha_r\log r + \beta_r)}_{\text{\normalfont image res. scaling}}}
\label{poly_eq}
\end{equation}
where $n$ is the amount of human-like video data used for self-supervised pretraining, $d$ is the model size (number of parameters), and $r$ is the image resolution (number of pixels). We thus fit 78 data points in each condition with the 6 parameters of this log-polynomial model. The Appendix contains the results with two alternative parametric scaling functions. The results with these alternative scaling functions are broadly consistent with the results presented here in the main text using the scaling function in Equation~\ref{poly_eq}.

\begin{figure}
  \centering
    \includegraphics[width=1.0\textwidth, trim=2mm 2mm 2mm 2mm, clip]{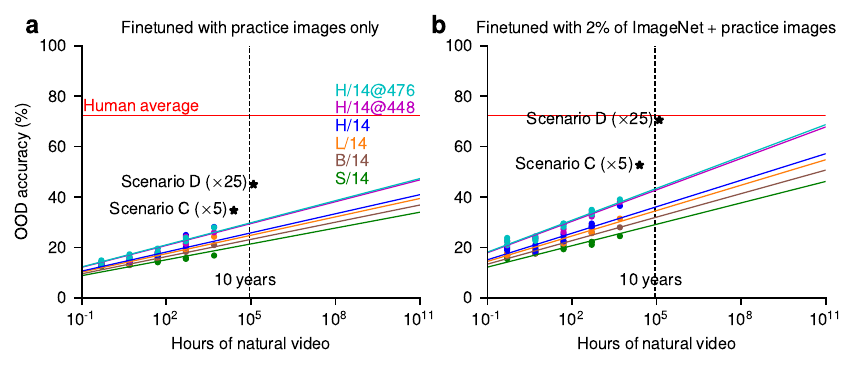}
  \caption{Results for the OOD ImageNet benchmark (figure description similar to Figure~\ref{imagenet_fig} above).}
  \label{ood_fig}
\end{figure}

\begin{table}
\centering
\begin{tabular}{cccccc}
 Scenario & $n$ (years) & $d$ (params) & $r$ (pixels) & Acc (stringent) & Acc (permissive) \\[1mm]
\hline
 Reference (actual) & 0.6 & 0.6B & 476$\times$476 & 28.0 & 39.0 \\
 Scenario C ($\times$5 projection) & 2.8 & 3.2B & 1064$\times$1064 & 34.7 & 52.7 \\
 \colorbox{lime}{Scenario D ($\times$25 projection)} & \colorbox{lime}{14.2} & \colorbox{lime}{15.8B} & \colorbox{lime}{2380$\times$2380} & 45.0 & \colorbox{lime}{70.6} \\
\end{tabular}
\caption{OOD accuracy on the OOD ImageNet benchmark under an actual reference scenario and two projections based on the fits to Equation~\ref{poly_eq}. The last two columns show the actual and projected accuracy under the stringent (finetuning with practice images only) and permissive (finetuning with 2\% of ImageNet + practice images) conditions respectively.}
\label{ood_tab}
\vspace{0em}
\end{table}

Based on these fits, Table~\ref{imagenet_tab} describes three scenarios: (i) an actual reference scenario that corresponds to the largest and most performant model we have trained, \textit{i.e.} a ViT-H/14 model trained with all $\sim$5K hours of our human-like video dataset with a spatial resolution of 476$\times$476 pixels; (ii) a hypothetical scenario A where we double ($\times$2) each of $n$, $d$, $r$ with respect to the reference scenario; (iii) a hypothetical scenario B where we quadruple ($\times$4) each of $n$, $d$, $r$ with respect to the reference scenario. Table~\ref{imagenet_tab} shows the actual and projected object recognition accuracy for each of these scenarios (actual accuracy for the reference scenario and projected accuracy based on the model fits for the hypothetical scenarios). In the hypothetical scenario B ($\times$4 projection) and under the more permissive finetuning condition (finetuning with 2\% of ImageNet), the projected accuracy already exceeds the 90\% top-5 accuracy we have set as the lower bound on human-level accuracy on ImageNet. This result is quite encouraging, since the increases in data size, model size, and image resolution required under this scenario are relatively modest: \textit{i.e.} this scenario only requires a 2.5B parameter ViT model trained with $\sim$20K hours (2.3 years) of human-like video with a spatial resolution of 952$\times$952 pixels.

Figure~\ref{ood_fig} shows the results of our scaling experiment on the OOD ImageNet benchmark. Similar to the case of ImageNet above, Table~\ref{ood_tab} describes an actual reference scenario and two hypothetical projections (a $\times$5 projection and a $\times$25 projection with respect to the reference scenario) based on the log-polynomial fits to the scaling data and shows the actual and projected OOD accuracies under these scenarios. In this case, we estimate that $n$, $d$, and $r$ need to be scaled up more compared to the case of ImageNet in order to reach human-level performance, but the $\times$25 scenario (scenario D) is projected to achieve human-level OOD accuracy under the permissive finetuning condition and, with its requirements of 14.2 years of data, a 15.8B parameter model, and a spatial resolution of 2380$\times$2380 pixels (5.7 MP), we believe this scenario is still broadly within the human bounds in all of $n$, $d$, and $r$.

\section{Discussion}
Our results suggest that human-level accuracy and robustness in visual object recognition are achievable from human-like visual experience at sub-human scales of data size, model size, and image resolution, using highly generic self-supervised learning algorithms and deep learning architectures without any strong inductive biases. This result is not inconsistent with our earlier work \citep{orhan2021}, which only considered the scaling of data size (fixing the model size and image resolution to relatively small values). In fact, in the current set of experiments, when we just consider the data size scaling for the ViT-S/14 model, which is similar in size to the ResNeXt-50 model trained with the DINO algorithm in \cite{orhan2021}, and fix the spatial resolution to 224$\times$224 pixels, we get estimates in the range of 1M-100M years of human-like visual experience in order to reach human-level accuracy on ImageNet (see the green lines in Figures~\ref{imagenet_fig}a and \ref{imagenet_fig}b). These estimates are comparable to the estimated range of 1M-1B years reported in \cite{orhan2021}.

\section*{Acknowledgements}
I would like to thank the authors of the MAE paper \citep{he2022} for making their code available. I would like to thank the creators and providers of the datasets used in this study. I would also like to thank the HPC team at NYU for their diligent maintenance efforts.
\bibliography{scaling}
\bibliographystyle{apalike}

\section*{Appendix}
\subsection*{Alternative scaling functions}

\begin{figure}
  \centering
    \includegraphics[width=1.0\textwidth, trim=2mm 2mm 2mm 2mm, clip]{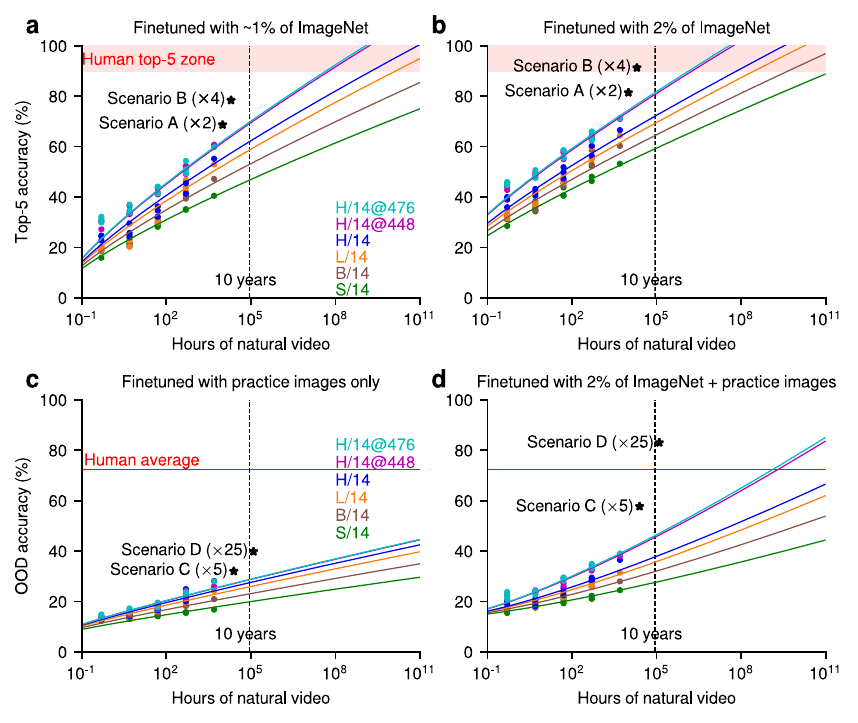}
  \caption{Results with the alternative scaling function in Equation~\ref{asf1_eq}. \textbf{a} and \textbf{b} show the results for the ImageNet benchmark; \textbf{c} and \textbf{d} show the results for the OOD ImageNet benchmark.}
  \label{asf1_fig}
\end{figure}

Figure~\ref{asf1_fig} shows the results with a 10-parameter scaling function of the form:
\begin{equation}
    \text{\normalfont accuracy} = (\alpha_n\Tilde{n}^{\beta_n} + \gamma_n)  (\alpha_d \Tilde{d}^{\beta_d} + \gamma_d)  (\alpha_r\Tilde{r}^{\beta_r} + \gamma_r) + \delta
\label{asf1_eq}
\end{equation}

\begin{figure}
  \centering
    \includegraphics[width=1.0\textwidth, trim=2mm 2mm 2mm 2mm, clip]{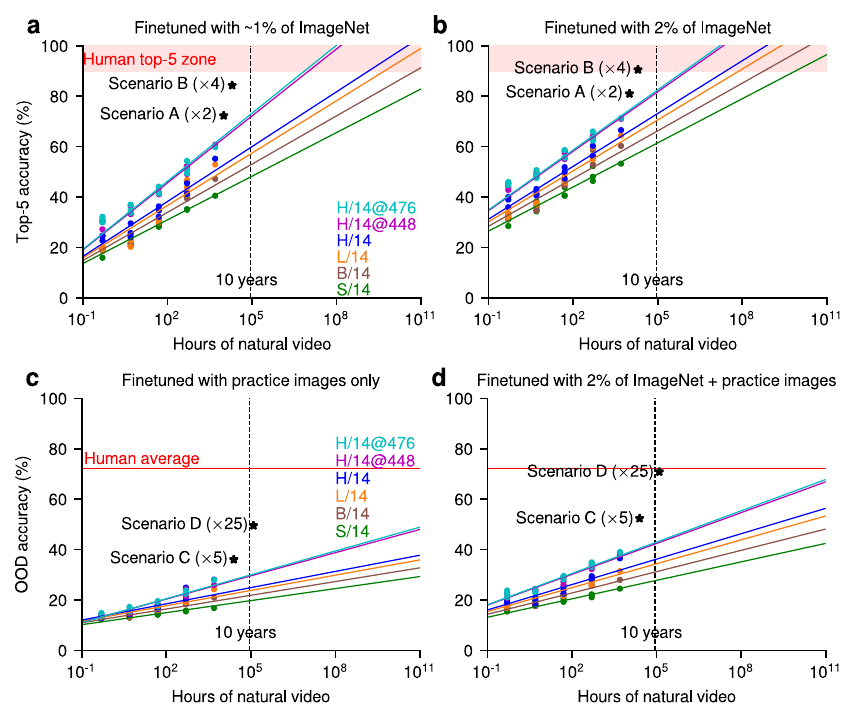}
  \caption{Results with the alternative scaling function in Equation~\ref{asf2_eq}. \textbf{a} and \textbf{b} show the results for the ImageNet benchmark; \textbf{c} and \textbf{d} show the results for the OOD ImageNet benchmark.}
  \label{asf2_fig}
\end{figure}

Figure~\ref{asf2_fig} shows the results with an 8-parameter scaling function of the form:
\begin{equation}
    \text{\normalfont accuracy} = \alpha_n \Tilde{n} + \alpha_d \Tilde{d} + \alpha_r \Tilde{r} + \alpha_{nd} \Tilde{n}\Tilde{d} + \alpha_{nr} \Tilde{n}\Tilde{r} + \alpha_{dr} \Tilde{d}\Tilde{r} + \alpha_{ndr} \Tilde{n}\Tilde{d}\Tilde{r} + \gamma   
\label{asf2_eq}
\end{equation}
where, in both equations, $\Tilde{n}$, $\Tilde{d}$, $\Tilde{r}$ represent $n$, $d$, $r$ in the log space: \textit{i.e.} $\Tilde{n}=\log n$, $\Tilde{d}=\log d$, and $\Tilde{r}=\log r$, respectively.

In Figures~\ref{asf1_fig}-\ref{asf2_fig}, panels \textbf{a} and \textbf{b} show the results for the ImageNet benchmark, panels \textbf{c} and \textbf{d} show the results for the OOD ImageNet benchmark under the stringent and permissive finetuning conditions, respectively. The reference scenario and the hypothetical scenarios considered here are identical to those in the main text. The scaling functions here and in the main text were fit with the \href{https://docs.scipy.org/doc/scipy/reference/generated/scipy.optimize.minimize.html}{\texttt{scipy.optimize.minimize}} function using the adaptive Nelder-Mead optimization method and the parameters were initialized to small values.

\end{document}